\documentclass[sigconf]{acmart}
\AtBeginDocument{%
  }

\begin{document}

\title{Beyond Physical Illness: Can Clinical Narratives Reveal Emerging Mental Health Risks in Chronically Ill Patients?}

%
\author{Sudeshna Jana}
\correspondingauthor
\affiliation{%
  \institution{TCS Research}
  \city{Kolkata}
  \country{India}
}
\email{sudeshna.jana@tcs.com}

\author{Siddhartha Laha}
\affiliation{%
  \institution{Jadavpur University}
  \city{Kolkata}
  \country{India}}
\email{siddharthalaha0@gmail.com}

\author{Manjira Sinha}
\affiliation{%
  \institution{TCS Research}
  \country{India}
}
\email{sinha.manjira@tcs.com}

\author{Tirthankar Dasgupta}
\affiliation{%
  \institution{TCS Research}
  \country{India}
}
\email{dasgupta.tirthankar@tcs.com}

\author{Pabitra Mitra}
\affiliation{%
  \institution{IIT Kharagpur}
  \country{India}}
\email{pabitra@cse.iitkgp.ac.in}

%
\renewcommand{\shortauthors}{Jana et al.}

%
\begin{abstract}
Mental-health risks such as depression, dementia, and suicidal ideation can emerge during hospitalization among chronically ill patients and may worsen without timely recognition and intervention. However, most existing mental health studies focus on patients with established psychiatric conditions rather than patients hospitalized primarily for physical illness. In this work, we investigate whether latent mental-health risk in chronically ill patients can be inferred from longitudinal clinical narratives. We construct a weakly supervised cohort around 4,000 patients from MIMIC-III using mental-health evidence extracted from discharge summaries. A hierarchical state-space framework derives mental-health representations from day-wise clinical narratives and models their temporal evolution for risk estimation throughout hospitalization. Preliminary results demonstrate improved binary risk prediction using the proposed representations and show distinct model derived risk trajectories between risk and no-risk cohorts, providing a foundation for studying emerging mental-health risks in chronically ill patients.

\end{abstract}




\maketitle

\section{Introduction}

Chronic illness can be accompanied by substantial mental health challenges, including depression, cognitive impairment, dementia, and self-harm risk. \cite{scott2023depression, akif2024impact} These conditions may emerge during hospitalization alongside the primary physical illness and can change rapidly in response to disease severity, medications, sleep disruption, and other clinical stressors. While some risks may be transient, others may persist beyond the acute hospitalization and have long consequences. Early recognition of such risks is therefore important for timely clinical intervention and patient safety. However, existing mental health research using clinical data has largely focused on populations with established psychiatric conditions, whereas the emergence of mental health risks among critically ill patients without a primary psychiatric diagnosis remains comparatively underexplored.\cite{nickson2023prediction, xu2023using}

Electronic health records (EHRs) provide rich longitudinal insights into a patient's hospitalization, encompassing clinical conditions, treatments, diagnoses, adverse events, and care plans.\cite{evans2016electronic} Nursing notes, in particular, are recorded throughout hospitalization and may routinely capture behavioral, cognitive, and affective signals associated with emerging mental-health concerns. However, identifying mental-health risk from these narratives presents two key challenges. First, these information is interwoven with extensive clinical information related to patients' medical history, medications, procedures, and physiological conditions, making it difficult to isolate mental-health specific signals. Second, explicit mental-health related documentation are typically unavailable, making direct supervised learning of day-wise risk difficult.

In this work, we address these challenges through a framework for inferring latent mental-health risk from longitudinal clinical narratives of ICU patients. The framework first constructs a weakly supervised cohort by extracting mental-health evidence from discharge summaries. Day-wise nursing narratives are then transformed into mental-health specific representations, which are subsequently modeled over time to estimate risk throughout hospitalization. This formulation enables investigation of not only whether a patient exhibits mental-health risk, but also how early such risk can be identified during the hospital stay.

The major contributions of this work are: (1) a weakly supervised cohort of approximately 4,000 ICU patients with evidence of mental-health risks, constructed from MIMIC-III; (2) a hierarchical state-space framework that learns mental-health-specific representations from day-wise clinical narratives; and (3) a temporal risk inference model that estimates mental-health risk throughout hospitalization and characterizes its model derived temporal trajectory.

\section{Cohort Construction}
As our primary data source, we use the publicly available MIMIC-III v1.4 dataset \cite{johnson2016mimic}, which contains records of approximately 40,000 critically ill patients. To construct the mental-health risk cohort, discharge summaries are used as the primary source of patient-level evidence, as they provide a comprehensive details of the hospitalization and consolidate relevant diagnoses, clinical events, patient history and discharge condition. Each discharge summary is processed using GPT-4o mini~\cite{zhang2024latest, reategui2025evaluation} with a structured prompt to identify explicit mental-health markers, extract their supporting evidence phrases, and determine their temporal context. Specifically, each identified marker is classified as either \textit{past history} or associated with the \textit{current hospitalization}, enabling us to distinguish pre-existing risks from those documented during the hospital stay.
The extracted evidence phrases are subsequently analyzed to identify the corresponding mental-health risks or conditions.
To reduce semantic redundancy among extracted markers, we employ an LLM-based concept-mapping agent coupled with the UMLS API~\cite{bodenreider2004unified}. For each extracted risk keyword, the agent retrieves candidate UMLS concepts and computes their cosine similarity with the keyword, selecting the highest-scoring concept as the standardized representation. This process consolidates semantically similar expressions, such as ``\textit{suicidal thoughts}'' and ``\textit{thoughts of killing himself}'', under a common clinical concept ``\textit{suicidal ideation}''.
The resulting patient-level labels retain the standardized concept, supporting evidence, and temporal context, providing weak supervision for subsequent mental-health risk modeling. We identified 4,113 patients with at least one of 143 unique mental-health risk markers.

\begin{table}[t]
\centering
\scriptsize
\setlength{\tabcolsep}{3pt}
\caption{Top five mental-health risks in the cohort, showing the number of patients with \textit{Past + Cur.} evidence in both prior history and current hospitalization, \textit{Cur.-only} evidence during the current hospitalization without identified prior history, and most frequent primary diagnoses.}
\label{tab:risk_stats}
\begin{tabular}{lccp{2.7cm}}
\toprule
\textbf{Risk} & \textbf{Past + Cur.} & \textbf{Cur.-only} & \textbf{Top primary diseases} \\
\midrule
Depression        & 616 & 892 & CHF; Sepsis \\
self-harm risk          & 356 & 524 & Carcinoma; Sepsis \\
hallucination           & 17 & 373 & Sepsis; Pneumonia \\
Dementia & 985 & 328 & Pneumonia; MI\\
Memory impairment         & 67 & 182 & MI; Sepsis \\
\bottomrule
\end{tabular}
\vspace{-15pt}
\end{table}

\subsection{Latent Mental-Health Risk Inference}

For each patient, nursing notes are documented at the day level to capture clinical evolution throughout hospitalization. We employ ClinicalMamba \cite{yang2024clinicalmamba}, a state-space sequence model designed for long clinical texts with selective information retention, to encode each day's narrative into a clinical representation capturing multiple aspects of the patient's condition, including physical and psychological information. Since these representations contain both mental-health related and broader clinical information, a shared projection followed by two parallel heads produces a mental-health representation $z_{\mathrm{m}}$ and a contextual representation $z_{\mathrm{c}}$. We jointly optimize the two representations using three complementary objectives. A class-weighted binary cross-entropy loss $\mathcal{L}_{\mathrm{m}}$ encourages $z_{\mathrm{m}}$ to capture information predictive of the identified mental-health risks. A reconstruction loss $\mathcal{L}_{\mathrm{r}}$ reconstructs the original clinical representation from $[z_{\mathrm{m}};z_{\mathrm{c}}]$, preserving clinically relevant information across the two representations. An orthogonality loss $\mathcal{L}_{\mathrm{o}}$ minimizes the cosine similarity between $z_{\mathrm{m}}$ and $z_{\mathrm{c}}$ for patients with identified mental-health risks, encouraging separation between mental-health specific and contextual information. The overall objective is
$\mathcal{L} =
\mathcal{L}_{\mathrm{m}}+
\lambda_{\mathrm{r}}\mathcal{L}_{\mathrm{r}}+
\lambda_{\mathrm{o}}\mathcal{L}_{\mathrm{o}}$,
where $\lambda_{\mathrm{r}}$ and $\lambda_{\mathrm{o}}$ control the relative contributions of the reconstruction and orthogonality objectives and are learned jointly with the model parameters.


The resulting sequence of daily mental-health representations is then processed by a temporal Transformer encoder~\cite{lim2021temporal} to model their evolution throughout hospitalization. The Transformer produces a risk probability for each hospitalization day and a final patient-level risk estimate. While the final prediction provides overall hospitalization risk, the day-wise probabilities characterize the temporal trajectory and allow the earliest model detected elevation in risk to be examined.

\section{Preliminary Results}
\begin{table}[t]
\centering
\scriptsize
\setlength{\tabcolsep}{4pt}
\caption{Binary mental-health risk prediction results.}
\label{tab:results}
\begin{tabular}{lccc}
\toprule
\textbf{Note Repres. + Temporal Model} & \textbf{F1} & \textbf{AUROC} & \textbf{AUPRC} \\
\midrule
Clinical Mamba + SSM              & 0.61 & 0.67 & 0.62 \\
Clinical Mamba + LSTM             & 0.63 & 0.75 & 0.71 \\
Clinical Mamba + Transformer      & 0.63 & 0.78 & 0.67 \\
Mental-health + SSM               & 0.71 & 0.82 & 0.82 \\
Mental-health + LSTM              & 0.75 & 0.85 & 0.82 \\
\textbf{Mental-health + Transformer}       & \textbf{0.78} & \textbf{0.86} & \textbf{0.85} \\
\bottomrule
\end{tabular}
\end{table}

\begin{figure}[t]
    \centering
    \includegraphics[width=0.95\columnwidth]{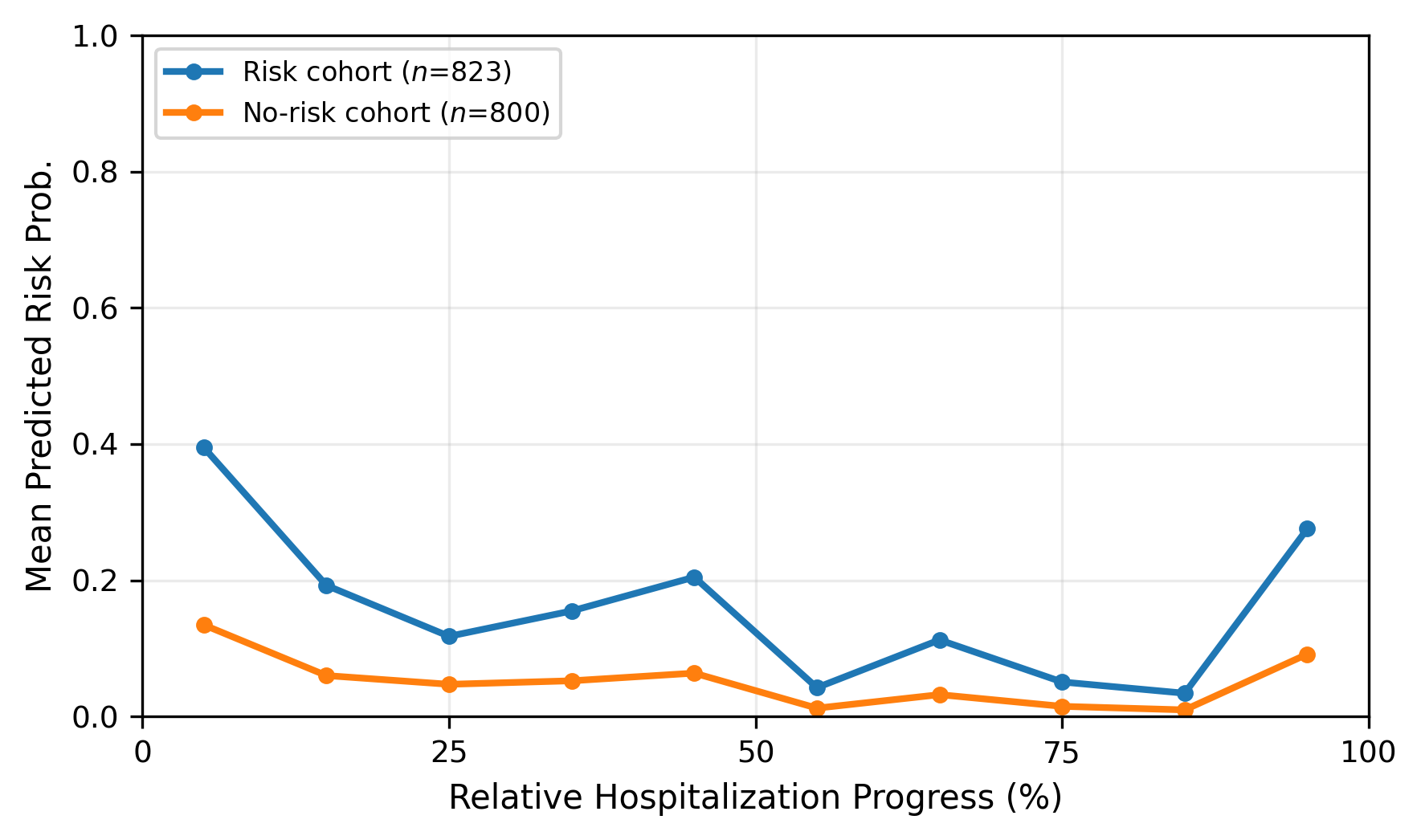}
    \vspace{-13pt}
    \caption{Predicted mental-health risk across relative hospitalization progress.}
    \label{fig:risk_trajectory}
\end{figure}

The experiments use 8,113 admissions, comprising 4,113 admissions with identified mental-health risk and 4,000 without risk. We evaluate binary patient-level risk prediction using the original Clinical Mamba representations and the proposed mental-health representations with three temporal architectures: SSM~\cite{rangapuram2018deep}, LSTM~\cite{elsworth2020time}, and Transformer~\cite{lim2021temporal}. As shown in Table~\ref{tab:results}, the mental-health representations consistently outperform the corresponding other representations across all temporal models.
This suggests that isolating mental-health specific information from the broader clinical representation provides a more informative basis for downstream risk prediction. 

In a further analysis, we examine the day-wise risk probabilities. To account for varying lengths of stay, predictions are normalized to relative hospitalization progress and averaged within intervals separately for the risk and no-risk cohorts. Figure~\ref{fig:risk_trajectory} presents the resulting model-derived risk trajectories, with the risk cohort generally exhibiting higher predicted risk throughout hospitalization.



\section{Conclusion}
This work presents a framework for inferring latent mental-health risk from longitudinal clinical narratives of chronically ill patients. Preliminary results show that mental-health-specific representations improve patient-level risk prediction and reveal distinct temporal risk trajectories between risk and non-risk cohorts. Future work will extend the framework beyond binary risk detection to identify specific mental-health risks and model their trajectories throughout hospitalization, capturing how patients' mental states emerge and evolve over time. We will further investigate interpretable approaches to identify the key clinical factors contributing to these changes, providing insight into not only \textit{what} mental-health risk emerges and \textit{when}, but also \textit{how} it evolves and \textit{why}.


\bibliographystyle{ACM-Reference-Format}
\bibliography{references}

@article{scott2023depression,
  title={Depression and anxiety trajectories in chronic disease: a systematic review and meta-analysis},
  author={Scott, Amelia J and Correa, Ashleigh B and Bisby, Madelyne A and Dear, Blake F},
  journal={Psychotherapy and Psychosomatics},
  volume={92},
  number={4},
  pages={227--242},
  year={2023}
}

@article{nickson2023prediction,
  title={Prediction and diagnosis of depression using machine learning with electronic health records data: a systematic review},
  author={Nickson, David and Meyer, Caroline and Walasek, Lukasz and Toro, Carla},
  journal={BMC medical informatics and decision making},
  volume={23},
  number={1},
  pages={271},
  year={2023},
  publisher={Springer}
}

@article{xu2023using,
  title={Using machine learning to predict antidepressant treatment outcome from electronic health records},
  author={Xu, Zhenxing and Vekaria, Veer and Wang, Fei and Cukor, Judith and Su, Chang and Adekkanattu, Prakash and Brandt, Pascal and Jiang, Guoqian and Kiefer, Richard C and Luo, Yuan and others},
  journal={Psychiatric Research and Clinical Practice},
  volume={5},
  number={4},
  pages={118--125},
  year={2023}
}

@article{evans2016electronic,
  title={Electronic health records: then, now, and in the future},
  author={Evans, R Scott},
  journal={Yearbook of medical informatics},
  volume={25},
  number={S 01},
  pages={S48--S61},
  year={2016},
  publisher={Georg Thieme Verlag KG}
}

@article{johnson2016mimic,
  title={MIMIC-III, a freely accessible critical care database},
  author={Johnson, Alistair EW and Pollard, Tom J and Shen, Lu and Lehman, Li-wei H and Feng, Mengling and Ghassemi, Mohammad and Moody, Benjamin and Szolovits, Peter and Anthony Celi, Leo and Mark, Roger G},
  journal={Scientific data},
  volume={3},
  number={1},
  pages={160035},
  year={2016},
  publisher={Nature Publishing Group}
}

@article{bodenreider2004unified,
  title={The unified medical language system (UMLS): integrating biomedical terminology},
  author={Bodenreider, Olivier},
  journal={Nucleic acids research},
  volume={32},
  number={suppl\_1},
  pages={D267--D270},
  year={2004},
  publisher={Oxford University Press}
}

@incollection{reategui2025evaluation,
  title={Evaluation of the performance of a large language model to extract signs and symptoms from clinical notes},
  author={Reategui-Rivera, C Mahony and Finkelstein, Joseph},
  booktitle={Envisioning the Future of Health Informatics and Digital Health},
  pages={71--75},
  year={2025},
  publisher={SAGE Publications 1 Oliver's Yard, 55 City Road, London, EC1Y 1SP}
}

@article{akif2024impact,
  title={The impact of chronic diseases on mental health: an overview and recommendations for care programs},
  author={Akif, Adnan and Qusar, MMA Shalahuddin and Islam, Md Rabiul},
  journal={Current psychiatry reports},
  volume={26},
  number={7},
  pages={394--404},
  year={2024},
  publisher={Springer}
}

@inproceedings{yang2024clinicalmamba,
  title={Clinicalmamba: A generative clinical language model on longitudinal clinical notes},
  author={Yang, Zhichao and Mitra, Avijit and Kwon, Sunjae and Yu, Hong},
  booktitle={Proceedings of the 6th Clinical Natural Language Processing Workshop},
  pages={54--63},
  year={2024}
}

@article{lim2021temporal,
  title={Temporal fusion transformers for interpretable multi-horizon time series forecasting},
  author={Lim, Bryan and Ar{\i}k, Sercan {\"O} and Loeff, Nicolas and Pfister, Tomas},
  journal={International journal of forecasting},
  volume={37},
  number={4},
  pages={1748--1764},
  year={2021},
  publisher={Elsevier}
}

@article{rangapuram2018deep,
  title={Deep state space models for time series forecasting},
  author={Rangapuram, Syama Sundar and Seeger, Matthias W and Gasthaus, Jan and Stella, Lorenzo and Wang, Yuyang and Januschowski, Tim},
  journal={Advances in neural information processing systems},
  volume={31},
  year={2018}
}

@article{elsworth2020time,
  title={Time series forecasting using LSTM networks: A symbolic approach},
  author={Elsworth, Steven and G{\"u}ttel, Stefan},
  journal={arXiv preprint arXiv:2003.05672},
  year={2020}
}

@article{zhang2024latest,
  title={The latest version ChatGPT powered by GPT-4o: what will it bring to the medical field?},
  author={Zhang, Nan and Sun, Zaijie and Xie, Yuchen and Wu, Haiyang and Li, Cheng},
  journal={International journal of surgery (London, England)},
  volume={110},
  number={9},
  pages={6018},
  year={2024}
}


\end{document}